\documentclass[11pt,a4paper]{llncs}
\usepackage{amsmath}
\usepackage{amssymb}
\usepackage{graphicx}
\usepackage{marvosym}
\usepackage{url}
\usepackage{fancyhdr}

\usepackage{geometry}
\newcommand{\keywords}[1]{\par\addvspace\baselineskip
\noindent\keywordname\enspace\ignorespaces#1}

\fancypagestyle{firstpage}{\fancyhf{}
}

\begin{document}

%\mainmatter  % start of an individual contribution

\title{\LARGE{Project SHADOW: Symbolic Higher-order Associative Deductive reasoning On Wikidata using LM probing}}

%\titlerunning{Lecture Notes in Computer Science: Authors' Instructions}

\author{\large{Hanna Abi Akl}}
\institute{\large{Data ScienceTech Institute, Université Côte d’Azur, Inria, CNRS, I3S,\\ Paris, France}}

%\author{Alfred Hofmann%
%\thanks{Please note that the LNCS Editorial assumes that all authors have used
%the western naming convention, with given names preceding surnames. This determines
%the structure of the names in the running heads and the author index.}%
%\and Ursula Barth\and Ingrid Haas\and Frank Holzwarth\and\\
%Anna Kramer\and Leonie Kunz\and Christine Rei\ss\and\\
%Nicole Sator\and Erika Siebert-Cole\and Peter Stra\ss er}

%\institute{Springer-Verlag, Computer Science Editorial,\\
%Tiergartenstr. 17, 69121 Heidelberg, Germany\\
%\url{http://www.springer.com/lncs}}


%\toctitle{Lecture Notes in Computer Science}
%\tocauthor{Authors' Instructions}

\maketitle

\thispagestyle{firstpage}

\begin{abstract}
We introduce SHADOW, a fine-tuned language model trained on an intermediate task using associative deductive reasoning, and measure its performance on a knowledge base construction task using Wikidata triple completion. We evaluate SHADOW on the LM-KBC 2024 challenge and show that it outperforms the baseline solution by 20\% with a F1 score of 68.72\%.
\end{abstract}

%\begin{abstract}
%The abstract should summarize the contents of the paper and should
%contain at least 70 and at most 150 words. It should be written using the
%\emph{abstract} environment.
%\keywords{We would like to encourage you to list your keywords within
%the abstract section}
%\end{abstract}

\section{Introduction}
Large language models (LLMs) have performed increasingly well in a wide range of semantic tasks including those involving leveraging knowledge from the models themselves \cite{wang2024large}. This lead to research avenues investigating the capabilities of these models in knowledge-related tasks involving knowledge graphs and ontologies on the one hand, and measuring the intrinsic knowledge contained in LLMs on the other hand \cite{wang2024large}.
The Language Model Knowledge Base Construction (LM-KBC\footnote{https://lm-kbc.github.io/challenge2024/}) challenge proposes to evaluate intrinsic language model (LM) knowledge using techniques like LM probing and prompting \cite{petroni2019language} to construct knowledge bases by completing triples of subject entities and relations with the relevant object entities.
In this work, we present SHADOW, or \textbf{S}ymbolic \textbf{H}igher-order
\textbf{A}ssociative \textbf{D}eductive reasoning \textbf{O}n \textbf{W}ikidata, a fine-tuned model on knowledge base triples, and evaluate it on the LM-KBC task. We follow a methodology inspired from associative deductive reasoning \cite{takano2010associative} and leverage that technique to incorporate it in re-defining the probing problem to train the model more effectively. Specifically, we aim to address the following research questions: Can LLMs use transfer learning as a means to leverage their generative abilities to understand an association task? Can they use deductive reasoning to solve it based on a neighboring task they have been previously trained on? The rest of the work is organized as follows. In section \ref{sec:related-work}, we discuss some of the related work. Section \ref{sec:experiments} describes our experimental framework. In section \ref{sec:results}, we report our results and discuss our findings. Finally, we conclude in section \ref{sec:conclusion}.

\section{Related work}
\label{sec:related-work}

LM probing has been studied and evaluated in different research avenues. In their work, \cite{vulic2020probing} study the information stored in LLMs with respect to their architecture, focusing on the factors behind their understanding of lexical semantics. 
Other techniques leverage prompting to encourage LLMs to use their knowledge more effectively to find better answers. \cite{alivanistos2022prompting} show that curating prompts manually and combining them with sets of entities make relatively small LLMs perform well on knowledge base construction. Similar work shows that LLMs can be
prompted to generate seemingly coherent responses to incoherent inputs\cite{cherepanova2024talking}.

Other lines of work treat LLMs as knowledge bases and employ query-based techniques to probe them for specific knowledge. \cite{petroni2019language} compare different transformer-based models by querying them with specific input knowledge and tracing whether this knowledge is retained in the LLMs. \cite{alkhamissi2022review} evaluate LLMs as knowledge bases using different techniques ranging from model editing to discrete prompting on defined metrics like interpretability and causal tracing.

Research has also shown work on LLMs as knowledge bases using external knowledge in the form of extended vocabulary \cite{loureiro2022probing}, knowledge sources like knowledge graph information \cite{he2024can} or by designing architectures that support external vectorized knowledge sources like Retrieval-Augmented-Generation (RAG) systems \cite{li2024language}. Finally, while deductive abilities for LLMs have been heavily investigated, \cite{cheng2024inductive} perform experiments to evaluate the inductive reasoning capabilities of these models and find that their deductive reasoning are much poorer than their inductive reasoning skills.

\section{Experiments}
\label{sec:experiments}

This section describes our experiments in terms of data, model and training process.

\subsection{Dataset}

The data provided by the organizers are triples of the form (subject, relation, object). The following relations are considered:
\begin{itemize}
    \item \textbf{countryLandBordersCountry:} Null values possible (e.g., Iceland)
    \item \textbf{personHasCityOfDeath:} Null values possible
    \item \textbf{seriesHasNumberOfEpisodes:} Object is numeric
    \item \textbf{awardWonBy:} Many objects per subject (e.g., 224 Physics Nobel prize winners)
    \item \textbf{companyTradesAtStockExchange:} Null values possible
\end{itemize}

The data is provided in 3 sets: train, validation and test. The test set is used as the official submission evaluation set. The number of triples in each set is:
\begin{itemize}
    \item 377 in the train set
    \item 378 in the validation set
    \item 378 in the test set
\end{itemize}

For the subject and object in every triple, both the ID and the label are provided. A sample triple is thus represented as such:
\textit{\{"SubjectEntity": "Belize", "SubjectEntityID": "Q242", "ObjectEntities": ["Guatemala", "Mexico"], "ObjectEntitiesID": ["Q774", "Q96"], "Relation": "countryLandBordersCountry"\}}.

\subsection{Model}

Formally, SHADOW is a generative model designed to solve the following function: $y = f(x)$, where $x$ is the input triple $(s, p, \cdot)$ missing object entities, $f$ is the function that generates a number in the set $\{1,2,3,4,5\}$ and $y$ is the corresponding template from the set of templates $T = \{t_1,t_2,t_3,t_4,t_5\}$, where $t$ is a SPARQL query responsible for completing the knowledge graph triple with the missing objects. We train SHADOW as a conditional generation model from a base flan-t5-small\footnote{https://huggingface.co/google/flan-t5-small} model and fine-tune it on the provided data. The training hyperparameters are configured as such: 
\begin{itemize}
    \item \textbf{learning\_rate:} 1e-04
    \item \textbf{train\_batch\_size:} 4
    \item \textbf{eval\_batch\_size:} 4
    \item \textbf{num\_epochs:} 20
    \item \textbf{question\_length:} 512
    \item \textbf{target\_length:} 512
    \item \textbf{lr\_scheduler\_type:} linear
    \item \textbf{optimizer:} Adam
    \item \textbf{betas:} 0.9, 0.999
    \item \textbf{epsilon:} 1e-08
\end{itemize}

\subsection{Setup}

We design our experiment to combine LLM probing with a symbolic component and indirectly evaluate the intrinsic knowledge found in LLMs on Wikidata knowledge graphs. We shift the focus away from generating correct SPARQL queries to retrieving the relevant objects for each subject and relation pair by designing templates containing the dynamic queries needed to answer the generic question: 

\textit{What Z completes the relationship Y for X?}, 

where \textit{X}, \textit{Y} and \textit{Z} refer respectively to the subject, relation and object(s) in a triple. Since the challenge deals with 5 types of relations, we design a total of 5 templates and assign a numerical template ID to each one of them. We then pair each subject and relation from a triple with the corresponding template ID which points to the correct SPARQL query that retrieves the corresponding object(s) to complete the triple based on the relation type. The SPARQL queries are themselves designed based on the Wikidata properties that best represent the targeted relation in the given triples.

\begin{figure}[htbp]
  \centering
  \includegraphics[width=\linewidth]{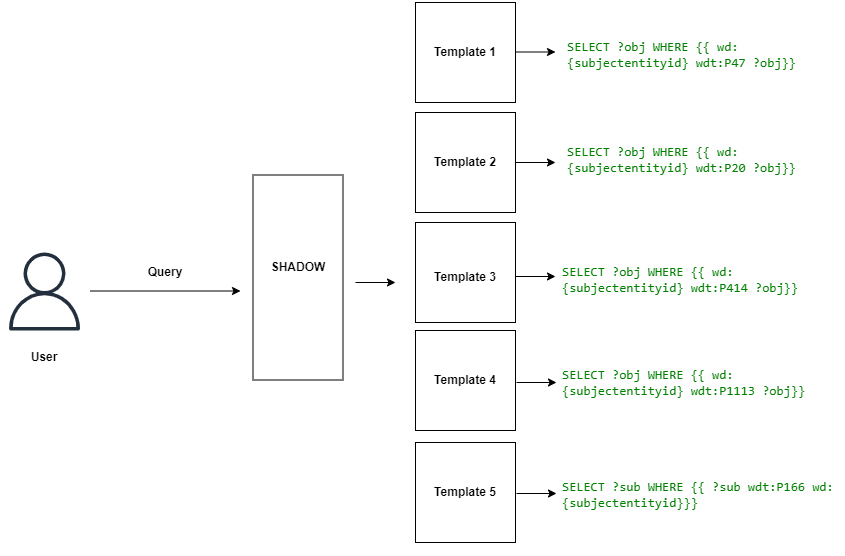}
  \caption{Experimental setup.}
\label{fig:setup}
\end{figure}

SHADOW is then trained to generate the correct template ID depending on the given subject and relation without seeing the SPARQL queries. This requires the model learning on two fronts: first is by implicitly learning to associate the correct template with the relation type, and second by learning to generate a numerical value which corresponds to one of the acceptable template IDs defined in the frame of the experiment. Figure \ref{fig:setup} shows our experimental design process. 

The training is done by splitting the train set into an 80-20 split randomly and training on the 80\%. The remaining 20\% are incorporated into the validation set. We use the scikit-learn\footnote{https://scikit-learn.org/stable/} library to perform the data splits. Table \ref{tab:training-results} shows the training results.

\begin{table}[htbp]
  \caption{SHADOW model training results}
  \begin{center}
  \begin{tabular}{|c|c|c|l|}
  \hline
    Training Loss&Epoch&Step&Validation Loss\\
    \hline
    0.6679 & 1.0& 1000& 0.0005\\
    0.3425&2.0&2000&0.0002\\
    0.2303&3.0&3000&0.0001\\
    0.1735&4.0&4000&0.000\\
    0.1394&5.0&5000&0.0001\\
    0.1167&6.0&6000&0.00009\\
    0.1006&7.0&7000&0.00008\\
    0.0882&8.0&8000&0.00007\\
    0.0785&9.0&9000&0.00006\\
    0.0707&10.0&10000&0.00006\\
    0.0643&11.0&11000&0.00005\\
    0.0590&12.0&12000&0.00005\\
    0.0545&13.0&13000&0.00004\\
    0.0506&14.0&14000&0.00004\\
    0.0473&15.0&15000&0.00004\\
    0.0443&16.0&16000&0.00003\\
    0.0417&17.0&17000&0.00003\\
    0.0394&18.0&18000&0.00003\\
    0.0374&19.0&19000&0.00003\\
    0.0355&20.0&20000&0.00003\\
  \hline
\end{tabular}
\label{tab:training-results}
\end{center}
\end{table}

The experiment is conducted on a Google Colab instance using a L4 High-RAM GPU. The code for our experimental setup is publicly available on GitHub\footnote{https://github.com/HannaAbiAkl/SHADOW}. The SHADOW model has been publicly released on Hugging Face\footnote{https://huggingface.co/HannaAbiAkl/shadow}.

\section{Results}
\label{sec:results}

Tables \ref{tab:per-relation-scores} and \ref{tab:zero-object-cases} capture the results on the official challenge test set. Overall, SHADOW performs well on the template identification task for the different relations. A closer inspection of the results shows that the model performs worse on the \textit{countryLandBordersCountry} relation which can be interpreted by the fact that the corresponding query targets property \textit{P47}, or \textit{shares border with}, which encompasses but is not limited to results sharing land borders (i.e. it also targets objects sharing sea borders). The nature of the property explains the high recall score, which retains a larger set of objects for that relation, and the low precision which reflects the few actual correct objects expected. The reason for choosing P47 is that it was the closest property that meets the required relation.

The result is validated in the zero-object cases, wherein the recall score is fairly high compared to a low precision score. This results is directly impacted by the choice of coding the queries as templates, sacrificing flexibility in results in favor of correct syntax instead of leaving the query generation in the hands of the model and risking some volatility in the results.

The relation \textit{seriesHasNumberOfEpisodes} shows a contrasting result. The tradeoff between perfect precision and zero recall suggests a cautious classification, whereby the net of positive results, i.e. relations correctly associated with the proper template ID, is small but accurate. The precision score is explained by the fact that the query behind the template targets the correct property and returns the correct object (which is either a number or null if there are no series episodes). The low recall means SHADOW did not learn to associate this particular relation with its correct template, drawing questions over the intrinsic knowledge for that type of relation in the model. It is also possible that the stark difference between the nature of this relation, which is the only one among the 5 to expect a purely numerical answer as opposed to Wikidata entities, has proven more challenging for SHADOW to learn despite the fine-tuning process it has undergone.

Finally, it is worth highlighting the model's great performance on the \textit{awardWonBy} relation, considering it is underrepresented at one-tenth of the other relations, i.e. 10 relations in the train, validation and test sets compared to approximately 100 for each of the four other relations. Since the model received less data for this relation compared to the others and still performed well, the performance could be explained by the nature of the internal knowledge the model has amassed for this relation, which in turn enables us to hypthesize on the good quality of the data given to this model at pre-training.
\begin{table}[htbp]
  \caption{Per-relation scores}
  \begin{center}
  \begin{tabular}{|c|c|c|l|}
    \hline
    Relation & Precision	 & Recall	 & F1-score\\
    \hline
    awardWonBy&0.9816&1.0000&0.9900\\
    companyTradesAtStockExchange&0.9950&1.0000&0.9971\\
    countryLandBordersCountry&0.7470&0.9717&0.7829\\
    personHasCityOfDeath&0.9700&1.0000&0.9700\\
    seriesHasNumberOfEpisodes&1.0000&0.0000&0.0000\\
    Average&0.9453&0.7297&0.6872\\
    \hline
  \end{tabular}
    \label{tab:per-relation-scores}
\end{center}
\end{table}

\begin{table}[htbp]
  \caption{Zero-object cases}
  \begin{center}
  \begin{tabular}{|c|c|l|}
    \hline
    Precision&Recall&F1-score \\
    \hline
    0.4975&0.90006&0.6408\\
    \hline
  \end{tabular}
  \label{tab:zero-object-cases}
  \end{center}
\end{table}

Table \ref{tab:leaderboard} shows the official performance of SHADOW compared to other solutions. Despite its difficulties with some relations, SHADOW performs very well with respect to the baseline and outperforms it by almost 20\%. It falls however short of the best scores and is outclassed by some margin.

\begin{table}
  \caption{Official submission leaderboard}
  \begin{center}
  \begin{tabular}{|c|l|}
    \hline
    Team Name&Average F1-score\\ 
    \hline
    davidebara&0.9224\\
    KB&0.9131\\
    RAGN4ROKS&0.9083\\
    WWWD&0.6977\\
    \textbf{DSTI}&\textbf{0.6872}\\
    NadeenFathallah&0.6529\\
    Rajaa&0.5662\\
    aunsiels&0.5076\\
    lm-kbc-organizer&0.4865\\
    \hline
  \end{tabular}
  \label{tab:leaderboard}
  \end{center}
\end{table}

\section{Conclusion}
\label{sec:conclusion}

In this work, we show how a fine-tuned LLM model can leverage intrinsic knowledge through LM probing and combine it with associative deductive reasoning to build disambiguated knowledge bases. The performance of SHADOW, our model, outperforms the baseline by disambiguating relation types and indirectly associating them with relevant knowledge graph completion queries. Our experiments show however that LLMs are highly influenced by the type of data they have already been trained on and possess uneven knowledge with respect to Wikidata relations which leaves much room for improvement in that area. Future work will focus on studying relation types in depth to improve LLM knowledge and using that knowledge to further evaluate LLM reasoning capabilities.

\vspace{2cm}

\section*{Authors}
\noindent {\bf Hanna Abi Akl} is a professor and researcher in logic, language, computation and neuro-symbolic artificial intelligence. He holds a Masters degree in Computer Science from the Université Côte d’Azur and an MSc in Data Science and Artificial Intelligence from Data ScienceTech Institute. His main areas of research are mapping language to symbolic structures, logic-based reasoning and graph-based knowledge representation methods. Hanna works as a researcher at INRIA and holds the Vice-Dean position at Data ScienceTech Institute.\\

\end{document}